\documentclass[conference]{csce}

\usepackage[hmargin=.75in,vmargin=1in]{geometry}
\usepackage[american]{babel}
\usepackage[T1]{fontenc}
\usepackage{times}
\usepackage{caption}

\usepackage{textcomp}
\usepackage{epsfig,graphicx}
\usepackage{xcolor}
\usepackage{amsfonts,amsmath,amssymb}
\usepackage{fixltx2e} % Fixing numbering problem when using figure/table* 
\usepackage{booktabs}

\title{\bf Feature Selection for Learning to Predict Outcomes of Compute Cluster Jobs with Application to Decision Support}     %%%% Replace with your title.

\author{\authorblockN{Okanlawon. Adedolapo$^1$, Yang. Huichen$^1$, Bose. Avishek$^2$, Hsu. William$^3$, Andresen. Dan$^4$, and  \\ Tanash. Mohammed$^4$}
\authorblockA{ Department of Computer Science, Kansas State University\\ Manhattan, Kansas, USA}}

\begin{document}

\maketitle                        %%%% To set Title and Author names.

\begin{abstract}%%%% Replace with your abstract.
We present a machine learning framework and a new test bed for data mining from the Slurm Workload Manager for high-performance computing (HPC) clusters. The focus was to find a method for selecting features to support decisions: helping users decide whether to resubmit failed jobs with boosted CPU and memory allocations or migrate them to a computing cloud. This task was cast as both supervised classification and regression learning, specifically, sequential problem solving suitable for reinforcement learning. Selecting relevant features can improve training accuracy, reduce training time, and produce a more comprehensible model, with an intelligent system that can explain predictions and inferences. We present a supervised learning model trained on a Simple Linux Utility for Resource Management (Slurm) data set of HPC jobs using three different techniques for selecting features: linear regression, lasso, and ridge regression. Our data set represented both HPC jobs that failed and those that succeeded, so our model was reliable, less likely to overfit, and generalizable. Our model achieved an R\textsuperscript{2} of 95\% with 99\% accuracy. We identified five predictors for both CPU and memory properties. 
\end{abstract}

\vspace{1em}

%%%%%%%%%%%%%%%%%%%%%%%%%%%%%%%%%%%%%%%%%%%%%%%%%%%%%%%%%%%%

\section{Introduction}
\noindent Predicting necessary resources to completely run a job requires appropriate information submitted at the beginning. We extended our earlier work [1] by designing a model trained on a recent, improved Slurm data set collected from the high-performance computing (HPC) cluster at Kansas State University. Slurm is a widely used system for cluster management and job scheduling for both large and small clusters [2]. Our supervised learning task explored three different regression analysis methods: linear regression, ridge regression, and lasso. With these methods, we derived additional features not available when jobs were submitted. In the past few years, using machine learning to estimate resources has become a growing interest. Matsunaga et al. identified the best machine learning algorithms for predicting execution time and resource requirements for jobs [3]. More recently, Mao et al. used a variant of the policy gradient method, REINFORCE, to train a neural network to optimize for such objectives as minimizing average job slowdown and completion time [4].  
\\
\indent Traditional approaches for scheduling jobs, backfilling for example, are largely based on workload heuristics that need a lot of domain knowledge and are less adaptable to changing workloads. Applying scheduling improvements, however, have been marred by failure to predict emerging parameters during execution. Our approach attempts to design robust methods that can be generalized across platforms [5]. Exploring job characteristics should make better decisions possible for users and administrators of these systems to specify resource estimates for their jobs. These characteristics can be collected throughout a job: when the job is submitted, when it is queued, during the job run, and after the job is either completed or failed. We also surveyed active users of our system to capture anything that could help model user behavior and improve system user habits. Combining all these characteristics can better reveal job scheduling patterns and help in designing better scheduling techniques to optimize system resources.  
\\
\indent Our paper contributes a comprehensive analysis of which features influence job prediction using a representative data set collected from historical log files of computer clusters. These features are mostly related to the CPU and memory information of the jobs. Additional features are related to the time required for jobs to run on the cluster as well as the number of cores.  We addressed the lack of adequate features available to accurately predict the resources required by these jobs. When users submit jobs, they provide information on the number of cores needed or memory requested, but this information is insufficient for predicting outcomes. Although much postmortem data can be gathered from historical log files after a job has been executed, this lack of predictive information at submission significantly reduces the level of confidence with which we can predict job outcomes. Our results rank the most useful predictive features; we used these features to predict entities important for both CPU and memory.

\indent Finally, we introduced ongoing work on a framework with a gradient boosting decision tree  based on LightGBM. This framework provides faster training speed and higher efficiency, lower memory usage, better accuracy, support for parallel and GPU learning, and the capability of handling large-scale data. Our goal was to derive the best prediction features using regression analysis. \\

\indent In the rest of the paper, we describe current and ongoing work, show how dour data set developed, describe our design architecture and implementation, and evaluate our results. 
 
\section{Machine Learning and Related Work}
\noindent Our interest in this work is born of recent growing interest in using machine learning to optimize HPC systems [3, 4]. While optimizing HPC systems was our goal, we focused on creating models to help users accurately specify  resources needed so their jobs will run to completion, particularly jobs related to the CPU and memory. Making these models agnostic can help with real-time monitoring of the state of the system and user requests across platforms [8]. We also used features, historical data, and survey data to model user behavior to give more insight into user behavior and possibly encourage better scheduling habits. Currently, users rely on heuristics only as well as their past experience to help them with scheduling. Administrators managing these systems have monitoring dashboards to help optimize system efficiency, but for large clusters, having administrators continually monitor individual jobs is impractical. This shows a need to better analyze specified user resource estimates for jobs.

\subsection{Feature Analysis} One important part of predictive modeling is to select the most useful features for prediction. Feature selection requires constructing a subset of input features x, where x belongs to X to determine the output variable Y. Removing irrelevant and redundant features improves accuracy, reduces overfitting, and saves computational time. We had to find the approximate relationship function f() between input X and output Y, but using all the features available from the data set are not needed as inputs, especially when the data set contains hundreds or thousands of features. In general, there are three feature selection methods: filters, wrappers, and embedded  [9].
\\

\indent {\bf Filter} methods use statistical measures to assign a score for each feature. The features were retained or removed from the data set using their ranking. The features are independent of the predictive model. Filter methods include the chi-square test, correlation coefficient, and the Fisher score.
\\
\indent {\bf Wrapper} methods train a model with different combinations of a subset of features, evaluating and comparing their importance. The model was iterated until the optimal subset was found[15]. The selection of features is on of several search problems, which can be roughly divided into exhaustive searches, heuristic searches, and random searches. Some common examples include forward selection, backward elimination, and recursive feature elimination.
\\
\indent {\bf Embedded} methods integrate feature selection as part of the model training process. This combines the characteristics of filter and wrapper methods and is mostly geared towards reducing overfitting.  The most common embedded feature selection method is regularization, with Lasso and Ridge regression as examples.
\\
\indent We used all three feature selection methods in pre-processing for our experiment. Five different selector algorithms from each of the three methods were chosen: Random Forest for feature importance as a filter feature selection method; Linear Regression, Lasso, and Ridge Regression as embedded feature selection methods; and Linear Regression as a wrapper selection method. All methods were implemented using {\tt scikit-learn} [10].

\subsection{Gradient Boosting}
Our feature selection focused on predicting two main features: MaxRSS and CPUTimeRaw. The former is the maximum resident set size of all tasks in a job while the latter is derived as the time used by a job or step in HH:MM:SS format. Table 1 shows a comprehensive list of all the features and their descriptions. 
\\
\indent For our classification tasks, we explored decision-tree algorithms in conjunction with boosting methods like adaptive boosting (AdaBoost) and LightGBM, a framework designed to make training faster and handle large amounts of data. With gradient boosting methods, we can train models gradually, additively, and sequentially. These methods are generally considered weak predictive models, but their performance can be boosted while training efficiently. 

\subsection{Graph Convolutional Networks}
Recently, the machine learning research community has focused on graph convolutional networks (GCNs) \cite{kipfGCN2017} as a way to use fewer labeled nodes in semi-supervised learning. In our application, this meant we used bootstrapping to predict outcomes for jobs from historical data from submitting users or related jobs as well as users in a heterogeneous information network. Traditional deep learning models like convolutional neural networks (CNNs) and recurrent neural networks (RNNs) treat each observation in the data set as independent and process the data set as a grid-based structure where the data set can be either text or image. The GCN, on the other hand, treats observations as having inter-relationships and works from a graph structured data set or a graph constructed from data sets. In fact, all grid based data sets can be considered a subset of graph structure data sets, but reverse is not usually valid. In graph representations of system data, each observation is represented as a node, and the relationships between the observations are represented by edges that connect nodes. GNCs are a very new approach but show outstanding performance; they are accurate, precise, and have good recall, in many use cases, outperforming both binary and multi-class classification analyses even if the data set is insufficiently annotated compared with CNN and RNN. A key challenge, however, was to determine good metrics to show relationships between distinct observations in our Slurm data set.

\subsection{GCN in HPC Analytics}
Reliable computing for users of HPC-clusters requires addressing the challenge of ensuring usersknow how to use HPC computing resources. Having sufficiently experienced computing cluster users is not guaranteed. Inexperience can lead to inaccurate estimates of resource needs. Job management platforms like Slurm do not require the user to provide all information directly in a job request. The Slurm data set does not provide all user specific information, we must explore and extract probable  implicit relationships in existing user information from the data set. If we use optimal user information from the Slurm data set, that can help us predict the future job status of a user and recommend required resources to the user before the job is submitted. 
Although much research on GCN has been conducted on domains like social network analysis \cite{socGCN2019}, natural language processing \cite{textGCN2019}, and signal processing, to the best of our knowledge, ours is the first research applying GCN to HPC data. To find relationships between observations in the HPC data set, we used some user information included in the Slurm data set and collected user  data from surveys taken when users submitted their jobs. Thus, we can connect observations or datarows from the data set. Surprisingly, in the Slurm data set, we were not limited to annotated data because each row in the data set represents a job submission event, and every row shows failure or success. Our goal was to create a graph from the Slurm data set and train a GCN model on this graph data set.

\begin{figure}[h]
\caption{Graph Structure of HPC data for GCN} 
\includegraphics[width=\linewidth]{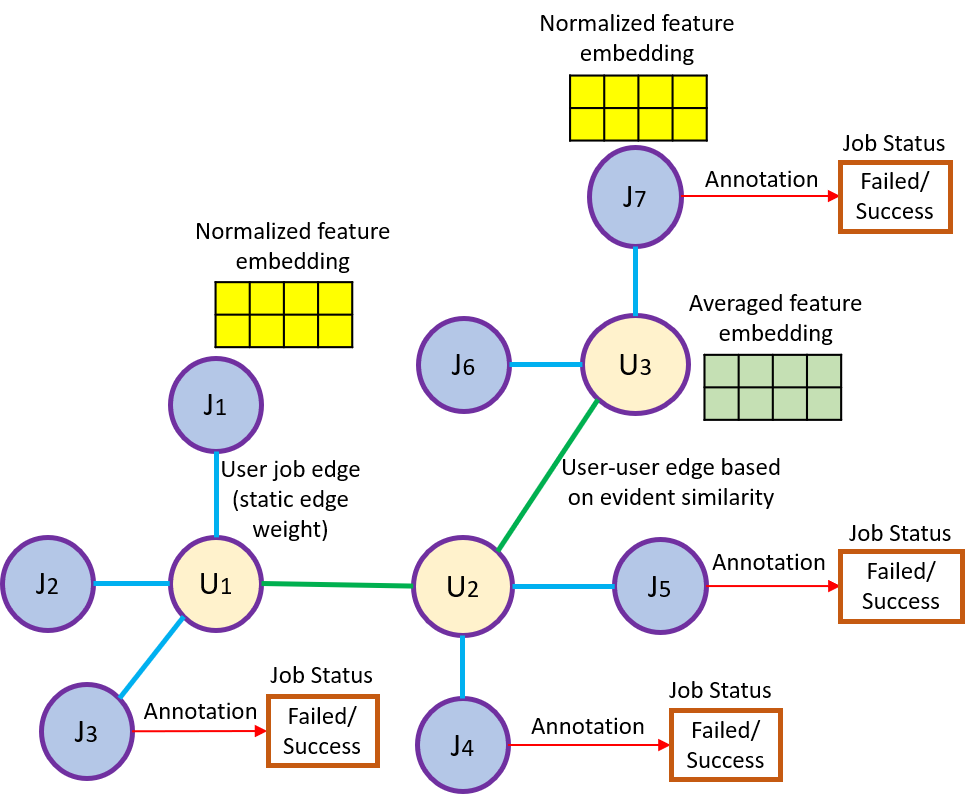}
\end{figure}

\subsection{Methods}
Figure 1 illustrates the modular structure of the network of user-job heterogeneous information. Label names for user nodes start with U while labels for job nodes start with J. Each is shown as a different color. User nodes are connected through edges derived from the relationships we extracted from the data set. Job nodes are not connected directly but through user nodes. We used the status of submitted job as the target for training the GCN model whether the job failed or succeeded. We extracted and then pre-processed some features from the Slurm data set. Afterward, we normalized the numeric features using min-max normalization and made an embedding from the datarow for each job submitted. We calculated the user node embedding by averaging all the embeddings of corresponding jobs submitted by the user. Finally, we trained a two layer GCN model on this graph data set. This approach is part of continuing work described in the conclusion.

\subsection{Stay-or-Go}
We derived insight from analyzing job-related and user-related information, but we also used scheduling decisions for jobs that were submitted and then failed for additional insight. The failure of HPC systems correlate to the type and intensity of the workload. Thus, when users decide to readjust job parameters during resubmission, this is also important. Therefore, we evaluated information obtained for jobs run on the local computing cluster both for one-time submissions and resubmissions after job failure. We compared this and information derived from migrating the job to a computing cloud like Amazon Web Services (AWS) [13] and Google Cloud Platform (GCP) [14]. With this information, user decisions are more informed, not just on how to set job parameters, but also to decide which platform is best for running their jobs and at what cost. \\
\indent Even though CPU and memory characteristics are important for predictive analytics of HPC jobs, we were also interested in the cost of a job that requires a number of cores on the system. We can define this cost by the length of time it takes for a job to run to completion if it was a one-time submission. For resubmitted jobs, this would be the time expended for each submission, as well as time required for a job to finish on the cluster. Another dimension is the dollar cost of running the job. For most on-premise clusters, the dollar cost is a flat fee even if the system is freely available as for a large number of our users. We can define a relationship between this on-premise cluster cost and the cost of migrating to the cloud, which can help users choose wisely. 

\indent Several studies have analyzed the performance of HPC applications on the cloud with interesting results[6, 7]. These studies show the viability of the cloud for HPC applications, with some concerns[8]. When the cloud is used as an alternative to running HPC jobs on local clusters, trade-offs in performance, interoperability, and security issues become concerns.

\subsection{Potential for Reinforcement Learning}
\noindent Reinforcement learning techniques could be used in the Stay-or-Go task of helping users decide when to stay on the local cluster or cloudburst (migrating to a cloud cluster, especially during peak periods) [17]. We model this as a sequential decision-making problem where an agent can be specified as a partially-observable Markov Decision Process (poMDP) [16].  In doing this, we can capture the uncertainty in decision-making, especially \textit{a priori} decisions made during multiple submissions of a job that continuously fails. This will help in optimizing the cost/utility function given below, so users can decide where to resubmit their jobs and what platform to use. \\

\[ \sum_i  C_i * Probability[(i, killed) | params_i] + C_A \] 

\noindent where C\textsubscript{i} is the cost of all runs up to time t, and C\textsubscript{A} is the extra cost of stay-or-go. This is evaluated in terms of time spent on either or both platforms and monetary costs of acquiring the cores to run the jobs.

\section{Experiment Design}
In this section, we describe acquiring and preparing training data for machine learning, the principles governing feature extraction and selection process, and design choices for learning algorithms and their parameters.

\subsection{Data Preparation}
The data set was collected from Simple Linux Utility for Resource Management (Slurm) database. Slurm represents the user job submission history from the year 2018 to 2019 of the primary Beocat HPC system with 10.9 million instances. The raw Slurm data set has 105 features that record parameters of any job that has finished running, including the number of resources that user requested when submitting the job and the number of resources the system allocated to the job. However, the raw data set has duplicated features, and some features have missing values. We cleaned the data set based on the following rules:

\begin{itemize}
	\item Remove meaningless features from the data set. For example, some features have only two values that are either very close to each other or one of which is unknown. The lack of variation in the features leads to unsuccessful regression results;
	\item Remove the features that have NaN values;
	\item Remove jobs that are running;
	\item Change all feature values to numeric values. For instance, change normal time format (Year-Month-Day, HH:MM:SS) to Linux timestamp.
\end{itemize}

\begin{table}
\caption{Features of the original data set}
\label{tab:xyz}
\begin{center}
\begin{tabular}{ |c|c|c|c| } 
 \hline
Account & JobName & AllocCPUS \\
AllocNodes & AveCPU & AveCPUFreq \\
AveRSS & CPUTimeRAW  & ElapsedRaw \\
Eligible & End & JobID \\ 
MaxDiskReadTask & MaxDiskWrite & MaxDiskWriteTask \\
MaxRSSTask & MaxVMSize & MaxVMSizeTask \\ 
NCPUS & NNodes & NTasks \\ 
ReqCPUS & ReqMem & ReqNodes \\ 
State & Submit & SystemCPU \\ 
Timelimit & TotalCPU & UserCPU \\ 
 \hline
\end{tabular}
\end{center}
\end{table}

\subsection{Design Architecture}
In this proposed approach both user and job will be represented as nodes. User-user relation will be made by considering various evident similarity metrics such as i) the project they are working, ii) the department they are from, iii) types of their submitted jobs, iv) job submission time, v) similarity of requesting resources, etc. We will select only those features from the data set which are important for class prediction by using feature selection algorithm. As the features' numeric values cover a wide range, we will do min-max normalization to make normalized feature embedding for job nodes. User node feature embedding will be calculated by averaging all submitted jobs by the user. We will split the data test and train set to train the model and predict the outcome whether the job will fail due to insufficient resources.

\begin{figure}[h]
\caption{Resource Estimation Pipeline} 
\includegraphics[width=\linewidth]{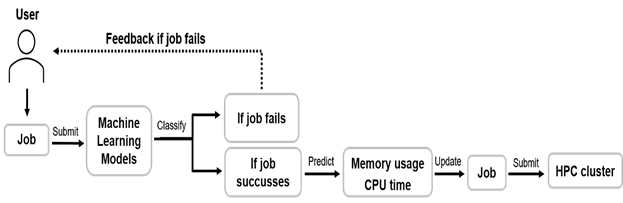}
\end{figure}

\subsection{Machine Learning Implementation}
To help improve prediction, we pre-processed our data set using the {\tt scikit-learn} [10] Python library before running our experiments, as documented in this section.

We chose CPUTimeRaw as the predicted feature for CPU, and mMaxRSS for memory. We created feature ranking scores that show the most useful features for prediction for three different methods: ;inear regression, Lasso, and Ridge Regression. We then aggregated these scores by calculating a mean score for our rankings. Tables 2 and 3 show the results for our top five features.

\begin{table}[htb]\centering
	\caption{Feature ranking scores for best features for predicting CPUTimeRaw}\label{t_sim}
	\begin{tabular}{@{}lccc@{}} \toprule
			Features & ranking score \\ \midrule 
	
			ElapsedRaw & 0.72 \\
			AllocCPUS & 0.66 \\
			NCPUS & 0.39 \\
			ReqCPUS & 0.36 \\
			Submit & 0.35 \\

			\bottomrule
		\end{tabular}
\end{table}

\begin{table}[htb]\centering
	\caption{Feature ranking scores for best features for predicting MaxRSS}\label{t_sim}
	\begin{tabular}{@{}lccc@{}} \toprule
			Features & ranking score \\ \midrule 
	
			AveRSS & 0.76 \\
			NNodes & 0.62 \\
			AllocNodes & 0.61 \\
			ReqMem & 0.54 \\
			Submit & 0.48 \\
			
			\bottomrule
		\end{tabular}
\end{table}

\section{Evaluation}
\noindent In this section, we discuss the rankings of predictive features using the mean score, which was derived using the means of the performance of all features across the three feature selection techniques. We also showed the performance of these predictors by applying linear regression on the data set and evaluating the predictive strength. This evaluated accuracy, F1, and AUC as metrics.
\\
Tables 2 and 3 show the best five features for prediction of the original 105 in the data set. This was calculated using the mean scores. Only AllocCPUS was highly predictive (70 percent) for the CPUTimeRaw while AveRSS and number of nodes (NNodes) performed better for MaxRSS. Table 4 shows the worst performing predictors for CPUTimeRaw are, as expected, features related to the memory. Similarly, Table 5 shows the worst predictors for MaxRSS.

\begin{table}[htb]\centering
	\caption{Feature ranking scores for worst features for predicting CPUTimeRaw}\label{t_sim}
	\begin{tabular}{@{}lccc@{}} \toprule
			Features & ranking score \\ \midrule 
	
			MaxRSS & 0.00 \\
			AveRSS & 0.01 \\
			AveCPUFreq & 0.02 \\
			NTasks & 0.03 \\
			MaxRSSTask & 0.04 \\
			
			\bottomrule
		\end{tabular}
\end{table}

\begin{table}[htb]\centering
	\caption{Feature ranking scores for worst features for predicting MaxRSS}\label{t_sim}
	\begin{tabular}{@{}lccc@{}} \toprule
			Features & ranking score \\ \midrule 
	
			MaxRSS & 0.00 \\
			AveRSS & 0.01 \\
			AveCPUFreq & 0.02 \\
			NTasks & 0.03 \\
			MaxRSSTask & 0.04 \\
			
			\bottomrule
		\end{tabular}
\end{table}

\indent The next step was running a linear regression model using features selected to predict MaxRSS and CPUTimeRaw. We also used k-fold cross validation with k = 5, selected empirically to reduce noise in the data.  We adopted the accuracy score library provided by scikit-learn to evaluate our models. Accuracy was the number of correct predictions made by our model divided by the total number of predictions made. F1 scores weighed both the precision and the recall, where an F1 score reached its best value at 1 and worst at 0. Finally, R-squared measured the percentage of variance explained by our models, the fraction by which the variance of the errors is less than variance of our predictors. Tables 5 and 6 show the results for both MaxRSS and CPUTime based on the metrics.

\begin{table}[htb]\centering
	\caption{Classification results for MaxRSS}\label{t_sim}
	\begin{tabular}{@{}lcccc@{}} \toprule  
			Model & Accuracy(\%) & F1(\%) & R squared (\%) \\ \midrule 

			LR  & 93 & 95 & 95\\ 
			
			LR with k-fold  & 98 & 96 & 95\\
			\bottomrule
		\end{tabular}
\end{table}

\begin{table}[htb]\centering
	\caption{Classification results for CPUTimeRaw}\label{t_sim}
	\begin{tabular}{@{}lcccc@{}} \toprule  
			Model & Accuracy(\%) & F1(\%) & R squared (\%) \\ \midrule 

			LR  & 96 & 91  & 93\\ 
			
			LR with k-fold  & 99 & 92 & 90\\
			
			\bottomrule
		\end{tabular}
\end{table}

\section{Conclusions and Future Work}
\noindent In this paper, we implemented a supervised learning model to extract the most useful features for predicting required HPC resources on a Slurm computing cluster. Our model achieved 99\% accuracy and confirmed the predictive ability of our model. Our model showed a few good predictors for scheduling. Those predictors should be further explored. 

\indent There is a potential for sequential-decision making formalization in this area especially by leveraging cloud resources. This could improve the ability to predict both time and computing resources needed to complete jobs.

\end{document}